# Graph-Enhanced Disease Trajectory Discovery and Temporal Endpoint Forecasting from Longitudinal Electronic Health Records

Jin-Jia Zheng[1], Vincent S. Tseng[1,*]

[1] Department of Computer Science, National Yang Ming Chiao Tung University, Taiwan

* Corresponding author: vtseng@cs.nycu.edu.tw

**Abstract**

Longitudinal electronic health records (EHRs) can reveal disease-progression structure, while useful forecasting requires leakage-controlled discovery, censor-aware outcomes, and inspectable trajectory evidence. We developed a graph-enhanced framework in MIMIC-IV v3.1 that splits patients before graph construction, learns train-only consecutive-admission transitions, and freezes endpoint-specific branching-motif policies selected on validation data. The final catalog contained 99,527 motifs, and a time-aware Transformer retrieved at most 24 patient-compatible motifs per endpoint through endpoint-specific cross-attention. Across five training seeds, micro-AUPRC increased from 0.9330 to 0.9357, while event-time MAE decreased from 676.9 to 657.7 days (reduction 19.2 days; 95% paired patient-cluster bootstrap CI, 13.8-24.8). Timing effects were heterogeneous: MAE worsened for events within 730 days but improved by 58.9 days for events occurring after 730 days. In an exploratory internal temporal split, branch conditioning reduced conditional event-time MAE by 26.9 days (95% CI, 23.0-30.8) without evidence of better micro-AUPRC. Corrected survival evaluation over the supported 0-2,555-day interval did not establish macro C-index or integrated Brier score superiority over a matched history-only discrete-hazard model. A directed, degree-matched GNN repair substantially improved unseen-edge discovery, yet replacing GNN-derived scores within fixed motif identities did not improve downstream forecasting. These results position branching motifs as structured trajectory context whose main value is long-range timing support rather than uniformly stronger endpoint discrimination, survival prediction, or causal explanation.

## 1. Introduction

Longitudinal EHRs contain repeated clinical encounters, diagnosis codes, irregular time intervals, and evolving comorbidity patterns [1]. In many clinical prediction settings, the objective is not only to determine whether a patient will eventually develop a target disease, but also to understand which prior conditions and disease transitions support that prediction and when the endpoint may occur. This is particularly important for chronic or progressive conditions such as chronic obstructive pulmonary disease (COPD), stroke, chronic kidney disease (CKD), and dementia, where disease histories often contain multiple overlapping signals rather than a single isolated precursor.

Neural sequence models have become a standard approach for EHR-based prediction. Recurrent models such as Doctor AI and GRU-D showed that time-stamped clinical histories can be encoded for future diagnosis or risk prediction [2, 3]. Attention-based models such as RETAIN and Dipole further improved interpretability by highlighting influential visits or code-level contributions [4, 5]. Transformer-based EHR models, including BEHRT, Med-BERT, Hi-BEHRT, and large generative forecasting models such as Delphi-2M, expanded the capacity to learn long-range disease dependencies and population-scale disease representations [6-10]. However, these models usually encode disease progression implicitly. Even when attention weights or token embeddings are inspected, it remains difficult to identify the explicit disease path or branching motif that supports a particular endpoint forecast.

A complementary line of work studies disease trajectories, transition graphs, chronological disease maps, and multimorbidity patterns [11-14]. These studies show that large-scale health records can reveal ordered disease-to-disease associations and population-level progression structure. Nevertheless, many trajectory-mining studies remain retrospective or descriptive: the discovered patterns are visualized, summarized, or used for hypothesis generation, but are not systematically integrated into endpoint ranking and event-time forecasting. A second limitation is that trajectory catalogs are often constrained by small curated sets or fixed top-N exports, which can arbitrarily restrict coverage and make it difficult to judge whether discovered evidence remains useful beyond a small subset of highly ranked motifs.

This paper addresses these gaps through a leakage-controlled graph-enhanced disease trajectory forecasting framework. Patients are split before discovery; the transition graph, GNN, support statistics, and motif candidates use train-only data; and stable versus broad catalog policies are selected by validation AUPRC and frozen before test evaluation. Patient-compatible branching motifs are encoded as branch tokens and combined with a time-aware Transformer to predict risk, at-risk endpoint ranking, and time to first recorded endpoint occurrence.

The paper addresses four aims: (1) whether train-only graph discovery can produce a high-coverage branching-motif catalog without a fixed top-N cap; (2) whether branch-token conditioning improves censor-aware risk, at-risk ranking, and event-time estimation; (3) whether the observed effects remain interpretable after candidate-count stratification, lead-time stratification, matched final-setting ablations, and a censor-aware survival comparator; and (4) whether improved unseen-edge discovery quality translates into incremental forecasting utility when motif identities are held fixed.

The main contributions are: (1) a patient-disjoint, train-only transition-graph and motif-discovery protocol; (2) a validation-selected endpoint-adaptive catalog of 99,527 branching motifs; (3) a censor-aware five-seed evaluation with 5,000 paired patient-cluster bootstrap resamples; (4) candidate-count-aware ranking, lead-time heterogeneity, matched ablations, calibration, and a tuned censor-aware hazard comparator; and (5) a directed-GNN repair study that separates unseen-edge discovery quality from downstream score utility, together with computational motif-deletion faithfulness analyses.

## 2. Related work

### 2.1 Longitudinal EHR forecasting

Longitudinal EHR forecasting has evolved from recurrent sequence models to increasingly large self-attention architectures. Doctor AI demonstrated that recurrent neural networks can learn representations of temporally ordered diagnosis and medication histories for next-visit prediction, while GRU-D explicitly incorporated informative missingness and irregular observation patterns into recurrent modeling [2, 3]. RETAIN and Dipole subsequently introduced attention mechanisms that make visit- or code-level contributions more inspectable without abandoning sequential prediction [4, 5]. These approaches established several design principles that remain central to clinical forecasting: the ordering of encounters matters, time gaps are informative, and the contribution of historical observations should be allowed to vary across prediction targets.

Transformer-based models extended this line of work by improving long-range dependency modeling and enabling large-scale pretraining. BEHRT adapts Transformer representations to longitudinal diagnoses and demographic context, Med-BERT uses large-scale structured EHR pretraining for disease prediction, and Hi-BEHRT extends hierarchical modeling to longer and multimodal patient histories [6-9]. More recently, Delphi-2M illustrates how generative Transformer modeling can learn population-scale disease histories and forecast multiple future conditions over long horizons [10]. Collectively, these studies show that powerful sequence encoders can summarize complex clinical histories and produce accurate future-event predictions across broad disease spaces.

However, most sequence models represent disease progression primarily through latent hidden states, attention weights, or token interactions. Such representations can indicate which prior visits or codes influence a prediction, but they do not necessarily expose an explicit, directional progression structure linking a patient's observed prefix to alternative future endpoints. They also tend to emphasize event probability more directly than the provenance of trajectory evidence used for risk ranking and event-time estimation. Our work is complementary to this sequence-modeling literature: the time-aware Transformer remains responsible for patient-state encoding, while graph-derived branching motifs provide an additional, inspectable source of endpoint-conditioned trajectory context. This separation allows us to test whether explicit progression structure adds information beyond a strong history-only encoder rather than replacing sequential EHR modeling.

### 2.2 Disease trajectory mining and transition graphs

Disease-trajectory research seeks to make longitudinal progression structure explicit rather than leaving it entirely inside a predictive representation. Population-scale studies have identified temporally ordered disease pairs and longer trajectories, chronological maps of hundreds of conditions, and interactive disease-progression browsers covering millions of patients [11-13]. These works demonstrate that directional co-occurrence patterns can be reproducibly extracted from longitudinal records and used to characterize common progression routes, disease clusters, and branching patterns at the population level. Related multimorbidity work has shown that latent disease combinations and recurring comorbidity structure can also be recovered from EHR data [14].

The objective of trajectory discovery is nevertheless different from patient-specific forecasting. Many population trajectory studies are primarily descriptive: they summarize statistically supported disease sequences, visualize transition structure, or provide hypothesis-generating maps [11-14]. Predictive trajectory models such as the Sequential Pattern Transformer move closer to forecasting by learning interpretable sequential patterns for future disease prediction [15], but the broader literature still leaves open how a large discovered trajectory catalog should be matched back to an individual patient's observed history and incorporated into endpoint-specific risk and timing models. This distinction is important because a population-level path can be statistically reliable yet irrelevant to a particular index admission, while a patient-compatible motif may carry useful context even when it is not among the globally most frequent patterns.

Catalog construction is therefore a methodological issue in its own right. Restricting analysis to a small top-ranked set can simplify interpretation but may reduce patient coverage and discard endpoint-specific structure; retaining a very broad catalog can preserve coverage at the cost of redundancy and computational burden. Our framework treats catalog policy as a validation-selected component rather than fixing an arbitrary global top-N threshold. In addition, it focuses on branching motifs that share an observed prefix and diverge toward alternative outcomes, because such motifs naturally correspond to the competing endpoint structure of the downstream forecasting task. This creates a direct bridge from population-level trajectory mining to patient-specific retrieval, endpoint ranking, and event-time estimation.

### 2.3 Graph-enhanced and knowledge-based clinical prediction

Graph-based clinical learning provides another route for injecting structured relationships into EHR models. GRAM uses a medical ontology to regularize code representations, G-BERT combines graph-based medical-code structure with Transformer pretraining, and heterogeneous clinical GNNs model relations among diagnoses, patients, or other EHR entities [16-19]. These applications build on general graph representation-learning principles introduced by graph convolutional networks, GraphSAGE, and graph attention networks [20-22]. Together, they show that clinical prediction can benefit from relational inductive biases that are difficult to express through a purely sequential representation.

At the same time, the type of graph and the way it enters the prediction model vary substantially across prior work. Ontology graphs encode curated semantic relations, patient-similarity graphs organize individuals by shared characteristics, and empirical transition graphs capture data-derived directional associations between diagnoses [16-19]. A latent graph embedding can improve predictive features without revealing which specific path or branch is active for a given patient. Conversely, an explicit transition path can be interpretable but may be unreliable if its edge evidence is sparse. These distinctions motivate separating graph representation quality, trajectory discovery, and downstream forecasting rather than assuming that improvement in one stage automatically transfers to the others.

Our approach uses the graph primarily as a discovery and scoring substrate. A directed link predictor contributes edge-level evidence, but the downstream model receives explicit patient-matched branching motifs rather than only node embeddings. This design makes it possible to test two related but distinct questions: whether graph-supported motifs help forecasting, and whether improving held-out edge prediction itself improves forecasts when motif identities are held fixed. The latter question is rarely isolated in graph-enhanced clinical prediction, yet it is important for understanding whether GNN scores function as essential predictive signals or mainly as reliability cues during discovery and prioritization.

### 2.4 Time-to-event and progression-sensitive evaluation

Clinical forecasting is incomplete if it considers only whether an endpoint occurs. Classical Cox proportional-hazards modeling provides a principled framework for censored time-to-event data, while neural extensions such as DeepSurv use nonlinear risk functions to retain a proportional-hazards objective [23, 24]. DeepHit and Dynamic-DeepHit further model event-time distributions and competing risks, with the latter incorporating longitudinal measurements for dynamic survival prediction [25, 26]. These methods emphasize the importance of censoring, risk-set construction, and event-time evaluation when clinical outcomes are only partially observed.

Longitudinal EHR endpoint forecasting differs from conventional single-outcome survival analysis in several ways. A patient can contribute multiple index admissions, several endpoints may remain simultaneously at risk, and diagnosis ascertainment is limited by the observation window of the source health system. For such settings, classification, endpoint ranking, and timing error answer complementary questions: discrimination measures whether future events are distinguished from non-events, ranking evaluates the relative ordering of currently eligible endpoints, and event-time metrics measure how closely predicted gaps match first recorded occurrences. A model can improve one dimension while remaining unchanged or even worsening on another, so aggregate classification performance alone can obscure clinically relevant behavior.

Evaluation design must also account for the dependence structure induced by repeated indices from the same patient and for heterogeneity across forecast horizons. Our study therefore uses authoritative censor-aware masks, evaluates timing only on observed positive at-risk events, and retains all repeated rows from a sampled patient within paired patient-cluster bootstrap resamples. We further stratify ranking by the number of eligible endpoints and timing by observed lead time, because ceiling effects and long-range versus near-term behavior can otherwise be hidden in overall means. This progression-sensitive evaluation connects the survival-analysis literature to the practical multi-endpoint structure of longitudinal EHR forecasting while avoiding a causal interpretation of recorded disease onset.

## 3. Proposed framework

Fig. 1 summarizes the leakage-controlled pipeline from patient-disjoint splitting and train-only discovery to validation-selected catalog freezing, patient-specific motif retrieval, and censor-aware test evaluation.

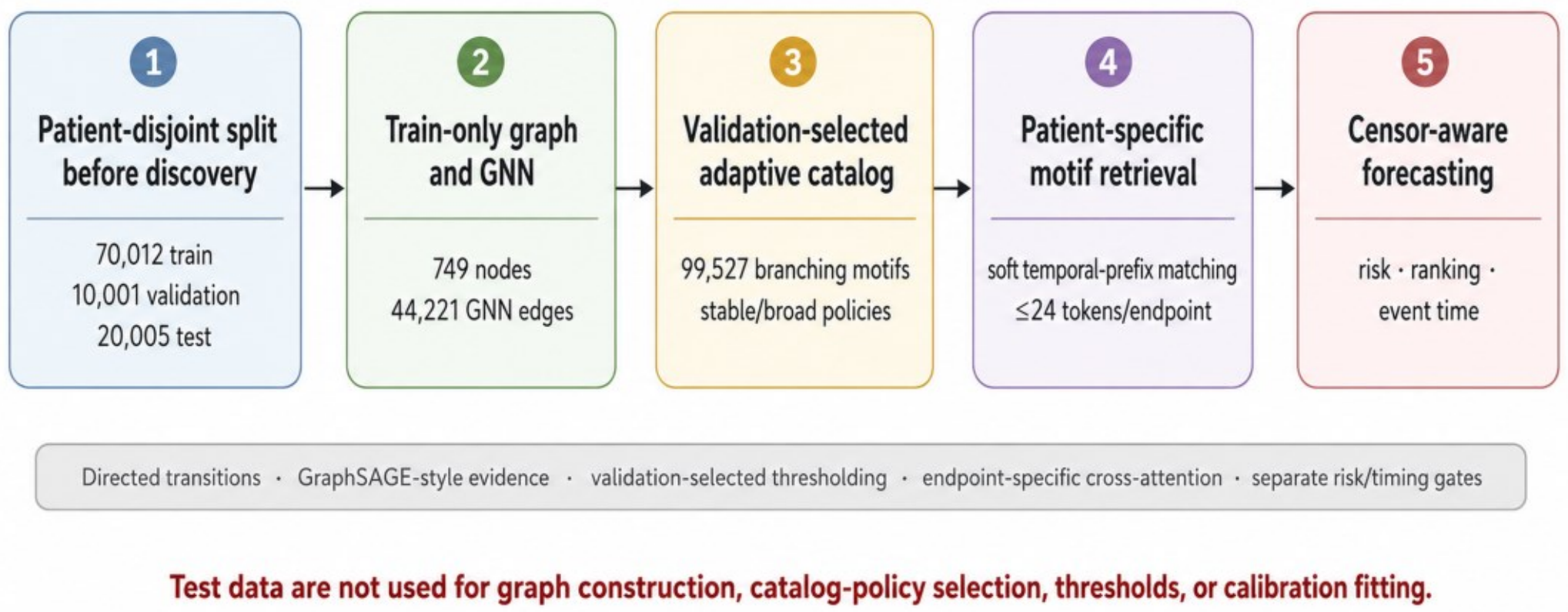

Fig. 1. Leakage-controlled graph-enhanced disease trajectory forecasting protocol. Patients are split before graph construction. Training data determine graph evidence and motif candidates; validation data select endpoint-specific stable or broad catalog policies; and the frozen 99,527-motif branching catalog is applied to censor-aware test examples. The final forecaster retrieves at most 24 motifs per endpoint and uses endpoint-specific cross-attention with separate risk and timing gates.

### 3.1 Problem setting, index examples, and endpoint outcomes

For patient i, each nonterminal admission j is a potential index admission with time $t_{ij}$. An endpoint observed at or before the index is prevalent and removed from the future at-risk set. Future occurrences within 30 days are washout exclusions. For horizon H, a row is positive when the first future recorded endpoint occurs from day 30 through H, negative when follow-up establishes no event through H, and unknown when no event is observed but endpoint-ascertainment follow-up is shorter than H. Unknown rows are excluded from loss and metrics.

$$F_{ij} = \max(0, t_{i,\text{last hospital}} - t_{ij}). \text{ For endpoint e, } y_{ije}(H)=1 \text{ when } 30 \leq T_{ije} \leq H;\ y_{ije}(H)=0 \text{ when } T_{ije}>H \text{ or no event is observed and } F_{ij}\geq H. \text{ The authoritative mask } m_{ije}(H)=0 \text{ for event-free rows with } F_{ij}<H \text{ and 1 otherwise.}$$

The outcome is multi-label, and one index may contain multiple future positive endpoints. Follow-up ends at the last recorded hospital discharge or an earlier recorded death; later post-discharge death does not extend diagnosis ascertainment. The term onset denotes the first recorded endpoint occurrence in MIMIC-IV rather than biological disease onset.

### 3.2 Consecutive-admission transition graph

Admissions were chronologically ordered for each patient. Directed diagnosis transitions were created only between consecutive admissions separated by 30 to 3,650 days. Each qualifying admission pair contributed the Cartesian product of unique mapped diagnoses in the earlier and later admissions, excluding self-transitions. Consequently, a given admission pair contributed at most one count to a particular directed diagnosis edge.

$$C_{uv} = \Sigma_{i,t}\, 1(u \in V_{i,t})\, 1(v \in V_{i,t+1})\, 1(30 \leq \Delta t_{i,t} \leq 3650), \quad u \neq v.$$
$$N_u = \Sigma_{i,t}\, 1(u \in V_{i,t})\, 1(30 \leq \Delta t_{i,t} \leq 3650), \quad \text{conf}_{uv} = C_{uv} / N_u.$$

Here $C_{uv}$ is transition support, $N_u$ is the number of qualifying source-admission opportunities containing u, and $\text{conf}_{uv}$ is empirical transition confidence. All graph quantities were constructed exclusively from the 70,012 training patients after a fixed subject-level split. The training graph contained 749 nodes, 228,000 observed directed edges before filtering, and 44,221 positive GNN training edges with support at least 50.

### 3.3 GraphSAGE-style directed link predictor

Each diagnosis node used six standardized structural features: log-transformed visit count, source count, target count, out-degree, in-degree, and mean outgoing transition confidence. The custom full-batch encoder used a symmetrized, row-normalized message-passing adjacency with self-loops, while the ordered edge scorer remained directed.

$$H^{(1)} = \text{ReLU}(XW_s^{(1)} + \tilde{A}XW_n^{(1)}), \quad H^{(2)} = H^{(1)}W_s^{(2)} + \tilde{A}H^{(1)}W_n^{(2)}, \quad Z_u = H_u^{(2)} / \|H_u^{(2)}\|_2.$$
$$\varphi(u,v) = [Z_u ; Z_v ; Z_u \odot Z_v ; |Z_u - Z_v|], \quad p^{gnn}_{uv} = \sigma(\text{MLP}_{256\rightarrow128\rightarrow1}(\varphi(u,v))).$$

The encoder dimensions were 6→128→64 with two message-passing layers and dropout 0.20. Positive edges were training-graph transitions with support at least 50; two directed non-edge negatives were sampled per positive. Weighted BCE used $w_{uv}=0.5+\log(1+C_{uv})/\log(1+C\max)$, AdamW for 100 epochs, learning rate 0.001, weight decay 0.0001, gradient clipping 5.0, and seed 42.

A post-primary discovery analysis evaluated directed incoming/outgoing message passing with target-degree-matched hard negatives, selected on validation data across seeds 42-46 and then locked for unseen-edge testing. This repaired GNN was supporting evidence only and did not replace the primary frozen catalog.

### 3.4 Explicit paths, branching motifs, and adaptive thresholds

The discovery code enumerates explicit two-hop paths A→B→C, one-to-many motifs $A\rightarrow\{B_1,\ldots,B_k\}$, and prefix-to-many motifs $A\rightarrow B\rightarrow\{C_1,\ldots,C_k\}$. Branch factor k is at most five and at least two. Explicit paths and branching motifs are both produced by discovery, but the final main forecaster is conditioned on branching motifs. The learned transition score combines graph probability, empirical support, and confidence:

$$T_{uv} = p^{gnn}_{uv} \log(1+C_{uv})\ conf_{uv}.$$

For a two-hop path, support is min(CAB,CBC), the GNN product is $p^{gnn}ABp^{gnn}BC$, and the ranking score is log(1+support) confAB confBC $p^{gnn}ABp^{gnn}BC$. For branching motifs, outgoing transition scores are normalized as $q_j = T_j/\Sigma_l T_l$ and branch entropy is $H = -\Sigma_j\, q_j \log q_j$. One-to-many and prefix-to-many ranking scores are:

$$R_1{\rightarrow}many = \log(1+\min_j C_AB_j)\ mean_j(T_AB_j)\ (1+H).$$

$$R_prefix{\rightarrow}many = \log(1+\min(C_AB, \min_j C_BC_j))\ T_AB\ mean_j(T_BC_j)\ (1+H).$$

The broad policy required branch support ≥50, mean branch GNN probability ≥0.80, and entropy ≥1.00; stable required ≥100, ≥0.90, and ≥1.20. Both disabled fixed top-N export. Validation AUPRC selected stable for COPD and dementia and broad for stroke and CKD. The frozen branching-only catalog contained 22,897 COPD, 15,828 stroke, 42,377 CKD, and 18,425 dementia motifs (99,527 total), comprising 832 one-to-many and 98,695 prefix-to-many motifs.

### 3.5 Soft temporal-prefix retrieval and branch-token representation

Motif matching is soft rather than a strict exact, subsequence, or suffix match. For a one-to-many motif with source A, match strength is 0.75 when A occurs in the index admission plus 0.25 when A occurs anywhere in earlier history. For a prefix-to-many motif $A \rightarrow B \rightarrow \{C_j\}$, match strength is the sum of 0.50 for A in the immediately preceding admission with B in the index admission, 0.25 for A anywhere in earlier history with B in the index admission, and 0.25 for B in the index admission.

$$m_one = 0.75 \cdot 1(A \in V_current) + 0.25 \cdot 1(A \in \text{earlier history}).$$

$$m_prefix = 0.50 \cdot 1(A \in V_previous, B \in V_current) + 0.25 \cdot 1(A \in \text{earlier history}, B \in V_current) + 0.25 \cdot 1(B \in V_current).$$

Candidate tokens were ranked by $r_j = m_j(0.5+G_j)(0.5+\tilde{R}_j)(1+0.1H_j)$, where $G_j$ is mean branch GNN probability and $\tilde{R}_j$ is the normalized log motif-ranking score within motif type. The frozen primary model used source-middle diverse-prefix retrieval and passed at most K=24 tokens to each endpoint query; K is a runtime input budget rather than a discovery-catalog size limit. Post-primary sensitivity analyses varied K from 4 to 48 and separately compared current-index versus historical-only matching and alternative diversity policies without changing the 99,527-motif catalog.

Each branch token combined source, optional middle, endpoint/target, and branch-type embeddings with a projected 14-dimensional numeric feature vector: match strength; one-to-many, prefix-to-many, current-source, current-middle, last-two-prefix, and any-history-prefix indicators; log support; normalized log ranking score; mean GNN probability; entropy; scaled median gap; temporal compatibility; and horizon-gap difference. The final main setting used branch_temporal_mode = none, so temporal-compatibility fields were retained in the schema but did not filter runtime retrieval.

### 3.6 Time-aware Transformer and endpoint-conditioned context gates

Each admission was encoded as a 749-dimensional multi-hot diagnosis vector fitted on the training vocabulary and projected to 128 dimensions. The log-normalized inter-admission gap passed through a 1→32→128 multilayer perceptron and was added to the visit embedding. A two-layer, four-head Transformer with feed-forward dimension 512, GELU activation, and dropout 0.20 encoded up to 24 admissions. The current implementation does not add a separate learned or sinusoidal absolute positional embedding.

For endpoint e, its embedding formed query $q_{ie}$ and attended to retrieved branch-token matrix $T_{ie}$ through four-head cross-attention. Risk and timing used separate cross-attention modules and separate scalar context gates:

$$c^{risk}_{ie} = MHA^{risk}(q_{ie},T_{ie},T_{ie}),\ \ g^{risk}_{ie} = \sigma(MLP([q_{ie};c^{risk}_{ie}])),\ \ z^{risk}_{ie} = q_{ie} + \beta^{risk}\, g^{risk}_{ie}\, c^{risk}_{ie}.$$

$$c^{time}_{ie} = MHA^{time}(q_{ie},T_{ie},T_{ie}),\ \ g^{time}_{ie} = \sigma(MLP([q_{ie};c^{time}_{ie}])),\ \ z^{time}_{ie} = q_{ie} + \beta^{time}\, g^{time}_{ie}\, c^{time}_{ie}.$$

The learnable β scalars were initialized to 0.25. These risk and timing context gates are the only gates used by the final main model. The optional learned relevance selector was disabled, its L1 and auxiliary loss weights were zero, no novelty file was supplied, and novelty auxiliary weight was zero. Accordingly, the final architecture contains neither a standalone relevance gate nor a novelty gate.

### 3.7 Multi-task objectives and optimization

The forecasting objective combined at-risk endpoint classification, observed-positive timing regression, and pairwise endpoint ranking:

$$\mathrm{L} = 1.0\ \mathrm{L_cls} + 1.0\ \mathrm{L_time} + 0.10\ \mathrm{L_rank}.$$

Classification used weighted binary cross-entropy over the endpoint-specific at-risk mask with positive element weight 8. Timing used SmoothL1 loss on normalized log gaps only for observed positive endpoints. Ranking used mean softplus($-s_i p+s_i n$) over at-risk positive–negative endpoint pairs in the same index example, with margin zero. Training used AdamW, learning rate 0.001, weight decay 0.0001, batch size 256, 40 epochs, gradient clipping at 5.0, and validation micro-AUPRC for checkpoint selection. Forecasting runs used seeds 42–46.

Table 1. Final configuration of the leakage-controlled graph-discovery and branch-token forecasting pipeline.

| Component | Final setting |
| --- | --- |
| Dataset | MIMIC-IV v3.1 |
| Patient split | 70:10:20 before graph construction |
| Transition graph | Train-only data; consecutive admissions; 30-3,650 days |
| Graph model | Primary 2-layer GraphSAGE-style link predictor; directed repair as supporting analysis |
| Catalog policy | Validation AUPRC; stable/broad thresholds; no fixed top-N cap |
| Frozen catalog | 99,527 branching motifs only |
| Runtime retrieval | Mapped-code soft prefix matching; diverse prefix; K=24 tokens/endpoint |
| Forecaster | 2-layer time-aware Transformer; 128 hidden; 4 heads; dropout 0.20 |
| Loss | BCE + 1.0 SmoothL1 gap + 0.10 pairwise rank |
| Evaluation | 5 seeds; 5,000 paired unique-patient bootstrap resamples |

## 4. Experimental setup

### 4.1 Data source, coding, endpoints, and cohort

Experiments used de-identified longitudinal diagnosis records from MIMIC-IV v3.1 [27, 28]. ICD-9-CM diagnoses were mapped to CCS categories and ICD-10-CM diagnoses to CCSR categories after removing decimal points and normalizing case. The final hybrid CCS/CCSR vocabulary excluded XXX000 and CCS259, retained all diagnosis rows without a seq_num restriction or maximum diagnoses per admission, and deduplicated mapped groups within each admission.

The source data contained 6,364,488 diagnosis rows, of which 6,258,931 were mapped. The eligible cohort contained 100,018 patients and 422,096 admissions. Patients were split 70:10:20 before discovery: 70,012/10,001/20,005 patients and 296,433/41,580/84,083 admissions in training/validation/test. Endpoint definitions were COPD (RSP008/CCS127), stroke (CIR020/021/022/024/025), CKD (GEN003/CCS158), and dementia (NVS011).

Validation selected the stable catalog for COPD and dementia and the broad catalog for stroke and CKD. Table 2 reports the locked cohort and final evaluation counts.

Table 2. Final leakage-controlled cohort, split, and evaluation counts.

| Scope | Patients | Admissions / indices | At-risk rows | Positive events |
|---|---|---|---|---|
| Train split | 70,012 | 296,433 admissions | - | - |
| Validation split | 10,001 | 41,580 admissions | - | - |
| Test split | 20,005 | 84,083 admissions | - | - |
| 3,650-day train | - | 56,054 indices | - | - |
| 3,650-day validation | - | 7,441 indices | - | - |
| Final test evaluation | 4,631 | 15,676 indices | 27,122 | 17,066 |
| Frozen catalog | - | 99,527 motifs | - | - |

### 4.2 Censor-aware outcome construction and follow-up

The index date was each nonterminal admission time. Endpoints recorded at or before index were prevalent and masked; future occurrences within 30 days were washout exclusions. Event-free rows with less than 3,650 days of ascertainment follow-up were unknown and excluded rather than treated as negatives. Follow-up ended at the last hospital discharge or an earlier recorded death.

The final outcome artifacts store authoritative endpoint masks, event times, and ascertainment follow-up. Timing errors are calculated only for observed positive at-risk events. This fixed-horizon censor-aware evaluation does not establish continuous observation outside the MIMIC hospital system and is not a discrete-time survival model.

### 4.3 Compared models

The principal matched comparison is the history-only time-aware Transformer versus the branch-enhanced model under the same split, censor masks, and checkpoint criterion. Additional same-task comparators included the SPT-inspired pattern-token adapter [15] and a randomly initialized Mamba-based history backbone [29], each using the same patient split, endpoint masks, visit/time inputs, endpoint heads, and training objectives. These are architecture-matched comparators rather than reproductions of the original authors' full tasks or pretrained pipelines.

Final-setting controls compared endpoint-conditioned cross-attention with mean-token aggregation, content-free null tokens that preserved token counts and masks, donor tokens drawn only from training patients, and removal of online numeric token channels. The matched no-branch ablation retained ranking-loss weight 0.10, permitting an isolated branch-token comparison. These controls were exploratory and did not replace the frozen primary model.

A supporting discrete-hazard analysis used a separate saved at-risk survival cohort and five fitted seeds; it did not replace the primary conditional event-time model. Corrected IPCW integrated Brier scores were evaluated only over the supported 0-2,555-day interval. History-hazard, branch-hazard, and DeepHit-style models were treated as survival extensions or comparators, not as the same estimand as observed-positive conditional event-time MAE.

An exploratory internal temporal validation used anchor-year groups (train 2008-2013, validation 2014-2016, test 2017-2022). The originally planned 3,650-day late-era horizon had too few negatives for meaningful discrimination; after a class-support audit, the reported temporal analysis used a 1,095-day horizon. The amendment inspected outcome-support counts, including test support, but did not use model scores to choose the horizon; it is therefore treated as exploratory rather than as fully test-blind external validation.

### 4.4 Metrics and reporting

Risk discrimination used micro AUROC/AUPRC over authoritative at-risk rows. Ranking excluded non-at-risk endpoints and was examined overall and among indices with at least two or three eligible candidates. Timing used MAE, RMSE, sMAPE, and WAPE on observed positive events, with lead-time strata of <=180, 181-730, and >730 days. The survival extension additionally reported endpoint-macro C-index and IPCW integrated Brier score over 0-2,555 days; its saved test universe was distinct from the 4,631-patient binary-forecasting test cohort.

Primary prediction comparisons are five-seed means with 5,000 paired unique-patient bootstrap resamples, retaining all repeated indices and endpoint rows for each sampled patient. Survival paired intervals used 1,000 patient-cluster resamples, re-estimating censoring weights within each resample. Training-seed SD and patient-bootstrap CI are reported separately. Overall outcomes were treated as major comparisons, whereas endpoint-specific, ablation, temporal, survival, and subgroup analyses were exploratory unless assigned to a named multiplicity family. Holm adjustment was primary and Benjamini-Hochberg adjustment secondary; percentile CIs are pointwise, and bootstrap-tail quantities are not exact permutation-test p-values.

## 5. Results

### 5.1 Discovery scale and branch coverage

The validation-selected adaptive catalog contained 99,527 endpoint-specific branching motifs: 42,377 CKD, 22,897 COPD, 18,425 dementia, and 15,828 stroke motifs. At least one motif was retrieved for 99.48% of frozen test examples, and 98.87% of positive endpoint rows had branch evidence. Coverage does not imply clinical validity or causal explanation.

### 5.2 Main forecasting results

The main five-seed comparison is reported in Table 3.

Table 3. Five-seed overall test performance and paired patient-cluster bootstrap intervals.

| Metric | History-only | Branch-enhanced | Difference | Relative | Paired 95% CI |
|---|---|---|---|---|---|
| Micro AUROC | 0.879443 | **0.883852** | +0.004409 | +0.50% | [0.002297, 0.006566] |
| Micro AUPRC | 0.932993 | **0.935698** | +0.002705 | +0.29% | [0.001389, 0.004120] |
| Masked MRR | 0.983420 | **0.984971** | +0.001551 | +0.16% | [0.000251, 0.002889] |
| Masked Hit@1 | 0.970726 | **0.973955** | +0.003228 | +0.33% | [0.000699, 0.005867] |
| Masked Hit@3 | 0.996288 | **0.996728** | +0.000440 | +0.04% | [-0.000263, 0.001195] |
| Gap MAE (d) | 676.94 | **657.71** | 19.23 lower | 2.84% lower | [13.83, 24.80] |
| Gap RMSE (d) | 889.77 | **856.28** | 33.49 lower | 3.76% lower | [27.54, 39.55] |
| Gap sMAPE | 72.61% | **70.05%** | 2.55 points lower | 3.52% lower | [2.09, 3.04] |
| Gap WAPE | 56.41% | **54.80%** | 1.60 points lower | 2.84% lower | [1.16, 2.05] |

Across five seeds, the branch-enhanced model improved micro-AUROC by 0.004409 (95% CI 0.002297-0.006566), micro-AUPRC by 0.002705 (0.001389-0.004120), masked MRR by 0.001551 (0.000251-0.002889), and masked Hit@1 by 0.003228 (0.000699-0.005867). Gap MAE decreased by 19.23 days (13.83-24.80), RMSE by 33.49 days (27.54-39.55), sMAPE by 2.55 percentage points (2.09-3.04), and WAPE by 1.60 points (1.16-2.05). Hit@3 remained ceiling-limited. The prespecified branch-versus-history AUPRC, MAE, and sMAPE comparisons each remained supported in Holm multiplicity sensitivity (adjusted paired-bootstrap tail probability 0.003599).

Overall discrimination gains were modest and endpoint-specific. Multiplicity sensitivity retained several COPD discrimination and timing effects and timing effects for CKD, dementia, and stroke, but not uniform endpoint-level superiority. The central interpretation therefore remains a modest overall gain concentrated in event-time estimation.

### 5.2.1 Candidate-count-aware endpoint ranking

Masked ranking had a structural ceiling because 73.56% of positive ranking examples contained one eligible endpoint. Among 3,604 examples with at least two candidates, MRR rose from 0.9373 to 0.9432 and Hit@1 from 0.8893 to 0.9015; NDCG and concordance also favored the branch-enhanced model. Unadjusted CIs were directionally consistent, but the eight candidate-count-by-metric comparisons did not remain below 0.05 after Holm adjustment.

### 5.2.2 Timing heterogeneity by observed lead time

Timing was strongly lead-time dependent. Branch conditioning increased MAE by 37.78 days for events within 180 days and by 36.85 days for events at 181-730 days, but reduced MAE by 58.87 days for events after 730 days (95% CI 51.25-66.56 days better). Thus, the aggregate timing benefit was driven by long-range progression timing rather than uniformly improved near-term estimation. Exploratory mitigation did not remove this trade-off: a short-gap prior and soft temporal compatibility produced overall MAEs of 655.14 and 656.29 days, respectively, while a 2:1 short/mid timing-loss reweight reduced <=180-day and 181-730-day MAE to 349.14 and 278.27 days but increased >730-day MAE to 960.78 days and overall MAE to 689.50 days. Under the same reweighted objective, branch conditioning remained 24.09 days better overall than reweighted history-only but was still worse by 25.10 and 25.74 days in the short and medium strata. A validation-selected timing-oriented catalog of 117,936 motifs changed overall MAE by only -1.15 days relative to the frozen primary catalog (95% CI -3.54 to 1.24).

### 5.2.3 Internal temporal validation

In the exploratory 1,095-day late-era evaluation, 4,753 test patients were assessed under the anchor-year split. Branch conditioning reduced conditional event-time MAE by 26.901 days (95% CI 23.047-30.759) and sMAPE by 3.066 percentage points (2.282-3.863). In contrast, the micro-AUPRC difference was -0.000614 (95% CI -0.008795 to 0.007324) and the micro-AUROC difference was -0.001757 (-0.006210 to 0.002241), so the analysis supports timing robustness but not improved or equivalent discrimination. The frozen-motif graph-edge recurrence proxy also fell from 99.48% in validation to 38.70% in test; this is a graph recurrence proxy rather than patient-level motif retrieval coverage.

## 5.3 Final-setting ablation

Table 4 reports the eight-variant, five-seed ablation under the same split, masks, training budget, and evaluation protocol. Removing branch tokens degraded the principal overall outcomes, and the adaptive catalog outperformed fixed top-300 per endpoint. After Holm adjustment within the exploratory ablation family, the full model remained supported over the no-branch variant for overall AUROC, AUPRC, MAE, and sMAPE, while the adaptive catalog remained supported over fixed top-300 for overall MAE and sMAPE. GNN scores and learned gates were supported mainly by timing; entropy retained small discrimination effects. The ranking-loss term was not essential and its removal slightly improved sMAPE. The no-entropy ablation removes the online explicit entropy factor/channel, not entropy from frozen catalog selection or stored ranking.

Table 4. Final-setting ablation under the locked leakage-controlled protocol.

| Variant | AUROC | AUPRC | MRR | Hit@1 | Gap MAE | sMAPE |
|---|---|---|---|---|---|---|
| **Full adaptive** | 0.883852 | 0.935698 | 0.984971 | 0.973955 | 657.71 | 70.05 |
| History-only | 0.879443 | 0.932993 | 0.983420 | 0.970726 | 676.94 | 72.61 |
| No branch; rank retained | 0.878988 | 0.933117 | 0.983730 | 0.971533 | 670.04 | 71.88 |
| No GNN score | 0.884535 | 0.936231 | 0.985073 | 0.974072 | 660.89 | 70.53 |
| No entropy | 0.882281 | 0.934862 | 0.984956 | 0.973852 | 658.60 | 70.12 |
| Identity gate | 0.883312 | 0.935770 | 0.985006 | 0.973955 | 663.93 | 70.62 |
| No ranking loss | 0.883637 | 0.935764 | 0.985075 | 0.974043 | 656.73 | 69.79 |
| Fixed top-300/ endpoint | 0.881449 | 0.934136 | 0.983398 | 0.970873 | 666.61 | 71.40 |

Supporting sensitivities kept the primary 99,527-motif catalog fixed unless explicitly stated. Current-index retrieval outperformed strict historical-only retrieval (micro-AUPRC +0.002275, 95% CI 0.001206-0.003334; MAE 8.55 days lower, 95% CI 5.09-11.89). Train-only transition-window variants from 0-3,650 to 30-1,825 days preserved the same broad pattern of modest discrimination gains and long-range timing benefit, while none eliminated short/medium degradation. Token-budget analyses from K=4 to 48 treated K as a runtime input bound rather than a discovery-catalog size cap.

### 5.3.1 Matched comparators and token controls

Across five seeds, the SPT-inspired adapter achieved micro-AUPRC 0.932283 and conditional event-time MAE 668.254 days, compared with 0.935698 and 657.712 days for the proposed model. The paired proposed-versus-SPT differences were +0.003415 in micro-AUPRC (95% CI 0.001374-0.005699) and 10.541 days lower MAE (5.715-15.536). A matched Mamba-based history comparator achieved micro-AUPRC 0.935391 and MAE 669.521 days; the proposed model was 11.809 days lower in MAE (95% CI 6.006-17.858), whereas its AUPRC advantage was only +0.000307 (95% CI -0.001911 to 0.002526).

The proposed model was also 11.983 days lower in MAE than mean-token aggregation and 13.504 days lower than training-patient donor tokens. Against content-free null tokens, however, the proposed-minus-null MAE difference was -3.512 days (95% CI -8.210 to 1.304), so an overall advantage over preserved token-count/mask structure was not established. A no-online-numeric-channel variant achieved 654.410 days versus 657.712 days for the proposed model (proposed-minus-ablation difference +3.302 days, 95% CI 0.151-6.358); this result did not remain supported after the scoped family adjustment and does not justify replacing the frozen primary model. Together, these controls support a narrower claim than necessity of motif content or every engineered numeric channel.

## 5.4 Calibration and survival sensitivity

Fig. 2 summarizes paired overall differences and final-setting timing ablations. Endpoint-specific results are reported in text as exploratory heterogeneity analyses, with family-wise multiplicity sensitivity summarized in Section 5.6.

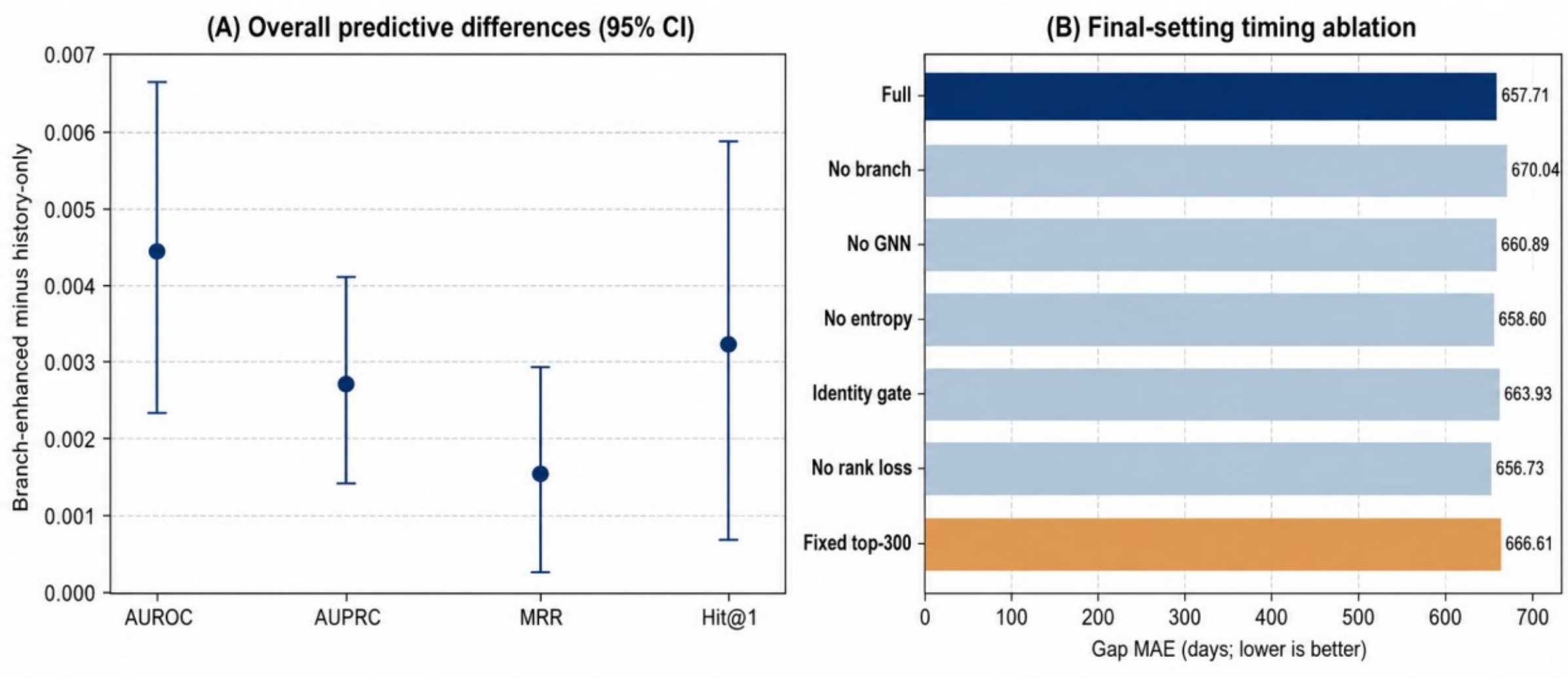

Fig. 2. Final leakage-controlled results. (A) Overall branch-enhanced minus history-only differences with 95% paired patient-cluster bootstrap intervals. (B) Gap MAE for final-setting ablations; lower values indicate better event-time estimation. The no-ranking-loss variant is retained as a transparent negative ablation rather than evidence that ranking loss is required.

Endpoint-specific discrimination gains remained concentrated in COPD. Validation-only Platt calibration reduced descriptive endpoint-by-seed mean Brier scores from 0.231757 to 0.137157 for history and from 0.222059 to 0.134445 for the proposed model; these are descriptive means rather than pooled micro estimates. In the corrected survival evaluation, history discrete hazard, DeepHit-style history, and branch discrete hazard achieved macro C-index/IBS[0,2,555 days] of 0.753974/0.140355, 0.750004/0.200132, and 0.751569/0.139957, respectively. Relative to matched history hazard, branch conditioning changed macro C-index by -0.002405 (95% CI -0.005166 to 0.000557) and IBS by -0.000398 (-0.001437 to 0.000577), establishing neither macro superiority nor equivalence. Endpoint effects were heterogeneous, including worse CKD IBS (+0.010966, 95% CI 0.009058-0.012766) and lower dementia concordance (-0.007282, -0.012290 to -0.002591). The substantially lower branch IBS versus the DeepHit-style comparator should not be interpreted as a full reproduction or universal survival advantage.

Because single-candidate indices force perfect ranks, Section 5.2.1 provides the primary interpretation of masked ranking; all-four-endpoint ranking is treated as sensitivity evidence.

### 5.5 Held-out graph and explanation evaluation

The original GNN had limited unseen-edge performance (AUPRC 0.1724; AUROC 0.5034). The validation-selected directed, degree-matched repair increased AUPRC to 0.3825 and AUROC to 0.7538, comparable to or better than the support-confidence heuristic for broad retrieval, although its MRR remained lower.

With all 99,527 motif identities and binary token availability fixed, replacing only GNN-derived scores produced no clear discrimination or ranking gain and increased MAE by 2.10 days (95% CI 0.35-3.86 days worse). Better unseen-edge prediction therefore did not translate into incremental fixed-catalog forecasting utility.

An additional matched graph sensitivity reinforced the separation between discovery quality and downstream utility. Adding a taxonomy-specificity feature increased degree-matched unseen-link AUPRC from 0.219982 to 0.356339, but forecasting MAE worsened from 656.445 to 660.168 days (increase 3.723 days; 95% CI 1.502-5.784). Generic filtering reduced unseen-link AUPRC to 0.176987 and worsened MAE by 4.150 days (95% CI 1.442-6.704). These graph extensions fixed GNN optimization seed 42; five link replicates varied negative sampling and are not five independent GNN training seeds.

Computational faithfulness was evaluated on frozen test examples. Deleting the top-attention motif changed endpoint probability by 0.0070 on average in absolute terms versus 0.0011 after random deletion and changed predicted event time by 21.6 versus 7.4 days. Deleting the top three motifs produced larger changes than deleting three random motifs. Signed effects were not uniformly positive, so attended motifs may be inhibitory as well as risk-increasing.

The final model retrieved at least one motif for 99.48% of test examples and provided branch evidence for 98.87% of positive endpoint rows.

### 5.6 Multiplicity sensitivity

Holm and Benjamini-Hochberg sensitivity analyses narrowed endpoint- and ablation-level claims but preserved the primary AUPRC and timing findings for branch conditioning and the adaptive catalog. These adjustments were applied to paired-bootstrap tail probabilities within prespecified families; percentile CIs remain unadjusted and are not simultaneous intervals.

## 6. Discussion

The leakage-controlled results clarify the role of graph-derived branching evidence. Branch tokens yield modest overall gains, but primary timing is heterogeneous: events within 730 days were estimated less accurately, whereas events beyond 730 days showed a 58.9-day MAE reduction. Exploratory internal temporal validation reproduced a conditional timing benefit without a discrimination gain, and timing priors or loss reweighting shifted error between short and long horizons rather than eliminating the trade-off.

Matched ablations and same-task comparators support the adaptive catalog and endpoint-conditioned token fusion while narrowing the mechanistic claim. The proposed model improved conditional timing relative to the SPT-inspired, Mamba-based, mean-token, and donor-token comparators, but did not establish an overall advantage over content-free null tokens, and the no-online-numeric-channel variant did not degrade. Extra capacity, token presence/count structure, and individual engineered numeric channels therefore cannot be ruled out or claimed as uniquely necessary.

Discovery quality and forecasting utility were separable: directed, degree-matched training repaired unseen-edge prediction, yet fixed-identity substitution of improved scores did not improve forecasts. A separate taxonomy-feature graph further improved unseen-link AUPRC while worsening downstream MAE. Branch-token value therefore does not increase monotonically with graph link-prediction quality.

Multiplicity sensitivity preserves the main branch-versus-history AUPRC and conditional timing conclusions while narrowing exploratory component claims. Corrected survival analysis did not establish macro C-index or IBS superiority over matched history hazard, so conditional event-time gains should not be generalized to survival prediction or clinical utility. Attention-guided motif deletion supports computational sensitivity, but not causality, clinical correctness, or clinician-validated reasoning.

## 7. Limitations and practical considerations

First, MIMIC-IV represents one hospital system, and the last recorded discharge does not guarantee continuous endpoint-free follow-up elsewhere. The censor-aware mask avoids treating known insufficient follow-up as negative, but single-system incomplete ascertainment remains.

Second, CCS/CCSR aggregation can merge heterogeneous diagnoses, and the transition graph represents observational association rather than causal disease mechanisms or biological onset. Endpoint-specific and ablation analyses are exploratory. Holm and Benjamini-Hochberg adjustments were applied to paired-bootstrap tail probabilities within prespecified families, but the reported percentile confidence intervals are not simultaneous multiplicity-adjusted intervals.

Third, ranking is ceiling-limited and candidate-count-stratified signals were not familywise significant after Holm adjustment. Fourth, timing worsened within 730 days and improved beyond 730 days; timing reweighting did not remove this short-versus-long trade-off. Fifth, null-token controls did not establish that motif content itself is necessary, and removal of online numeric channels did not impair the frozen setting. Sixth, repaired or taxonomy-augmented GNNs improved link prediction without improving downstream forecasting.

Seventh, the corrected survival extension relies on an independent-censoring interpretation and a supported IBS interval of 0-2,555 days, uses a different eligible test universe from the binary forecasting analysis, and did not establish macro superiority over history hazard. Eighth, the Mamba and DeepHit-style comparators are task-matched adapters rather than complete reproductions of pretrained external pipelines. Explanation analyses measure computational sensitivity rather than causal effect, clinical correctness, or clinician agreement. Internal temporal validation is not external-hospital validation; external longitudinal EHR validation, independent blinded clinician review, and institution-specific deployment evaluation remain future work.

The model is intended for retrospective methodological research and hypothesis prioritization, not direct clinical decision-making. Code, configurations, non-identifiable summary tables, and figure-source data are planned for release subject to institutional review and the MIMIC-IV data-use agreement.

## 8. Conclusion

This study presents a leakage-controlled graph-enhanced temporal endpoint forecasting framework. A frozen 99,527-motif catalog and endpoint-specific branch-token fusion produced modest overall gains and reduced aggregate event-time MAE by 19.2 days versus a matched history-only Transformer. Candidate-count analysis showed a directionally consistent but multiplicity-limited ranking signal, and lead-time stratification showed that timing gains were driven by events beyond 730 days. Same-task SPT-inspired and Mamba-based comparators and an internal temporal split supported a conditional timing advantage, but null-token and no-online-numeric controls narrowed claims about which motif components are necessary. Directed GNN improvements did not translate monotonically to forecasting, and corrected survival analysis did not establish macro superiority over matched history hazard. Branching motifs are therefore best viewed as inspectable structured context for retrospective conditional forecasting, not as causal explanations, guaranteed survival improvements, or evidence that stronger graph link prediction automatically yields stronger clinical forecasts.

## References

[1] Yadav, P., Steinbach, M., Kumar, V., & Simon, G. (2018). Mining electronic health records (EHRs): A survey. *ACM Computing Surveys, 50*(6), Article 85. https://doi.org/10.1145/3127881.

[2] Choi, E., Bahadori, M. T., Schuetz, A., Stewart, W. F., & Sun, J. (2016, December). Doctor AI: predicting clinical events via recurrent neural networks. In *Machine Learning for Healthcare Conference* (pp. 301-318). PMLR.

[3] Che, Z., Purushotham, S., Cho, K., Sontag, D., & Liu, Y. (2018). Recurrent neural networks for multivariate time series with missing values. *Scientific Reports, 8*(1), 6085.

[4] Choi, E., Bahadori, M. T., Sun, J., Kulas, J., Schuetz, A., & Stewart, W. (2016). RETAIN: An interpretable predictive model for healthcare using reverse time attention mechanism. *Advances in Neural Information Processing Systems*, 29.

[5] Ma, F., Chitta, R., Zhou, J., You, Q., Sun, T., & Gao, J. (2017, August). Dipole: diagnosis prediction in healthcare via attention-based bidirectional recurrent neural networks. In *Proceedings of the 23rd ACM SIGKDD International Conference on Knowledge Discovery and Data Mining* (pp. 1903-1911).

[6] Vaswani, A., Shazeer, N., Parmar, N., Uszkoreit, J., Jones, L., Gomez, A. N., ... & Polosukhin, I. (2017). Attention is all you need. *Advances in Neural Information Processing Systems*, 30.

[7] Li, Y., Rao, S., Solares, J. R. A., Hassaine, A., Ramakrishnan, R., Canoy, D., ... & Salimi-Khorshidi, G. (2020). BEHRT: Transformer for electronic health records. *Scientific Reports, 10*, 7155. https://doi.org/10.1038/s41598-020-62922-y.

[8] Rasmy, L., Xiang, Y., Xie, Z., Tao, C., & Zhi, D. (2021). Med-BERT: pretrained contextualized embeddings on large-scale structured electronic health records for disease prediction. *npj Digital Medicine, 4*, 86. https://doi.org/10.1038/s41746-021-00455-y.

[9] Li, Y., Mamouei, M., Salimi-Khorshidi, G., Rao, S., Hassaine, A., Canoy, D., ... & Rahimi, K. (2023). Hi-BEHRT: hierarchical transformer-based model for accurate prediction of clinical events using multimodal

longitudinal electronic health records. *IEEE Journal of Biomedical and Health Informatics, 27*(2), 1106-1117. https://doi.org/10.1109/JBHI.2022.3224727.

[10] Shmatko, A., Jung, A. W., Gaurav, K., Brunak, S., Mortensen, L. H., Birney, E., ... & Gerstung, M. (2025). Learning the natural history of human disease with generative transformers. *Nature, 647*, 248-256. https://doi.org/10.1038/s41586-025-09529-3.

[11] Jensen, A. B., Moseley, P. L., Oprea, T. I., Ellesøe, S. G., Eriksson, R., Schmock, H., ... & Brunak, S. (2014). Temporal disease trajectories condensed from population-wide registry data covering 6.2 million patients. *Nature Communications, 5*, 4022.

[12] Kuan, V., Denaxas, S., Gonzalez-Izquierdo, A., Direk, K., Bhatti, O., Husain, S., ... & Hingorani, A. D. (2019). A chronological map of 308 physical and mental health conditions from 4 million individuals in the English National Health Service. *The Lancet Digital Health, 1*(2), e63-e77.

[13] Siggaard, T., Reguant, R., Jørgensen, I. F., Haue, A. D., Lademann, M., Aguayo-Orozco, A., ... & Brunak, S. (2020). Disease trajectory browser for exploring temporal, population-wide disease progression patterns in 7.2 million Danish patients. *Nature Communications, 11*(1), 4952.

[14] Hassaine, A., Canoy, D., Solares, J. R. A., Zhu, Y., Rao, S., Li, Y., ... & Salimi-Khorshidi, G. (2020). Learning multimorbidity patterns from electronic health records using non-negative matrix factorisation. *Journal of Biomedical Informatics, 112*, 103606.

[15] Assadi Shalmani, M., Khani, M., Taleban, A., Yi, Z., Fink, J. T., Weber, C. E., Lu, Q., & Luo, J. (2026). Sequential pattern transformer (SPT): a generative and interpretable framework for predicting disease trajectories. *Neural Computing and Applications, 38*, 28. https://doi.org/10.1007/s00521-025-11695-4.

[16] Choi, E., Bahadori, M. T., Song, L., Stewart, W. F., & Sun, J. (2017, August). GRAM: graph-based attention model for healthcare representation learning. In *Proceedings of the 23rd ACM SIGKDD International Conference on Knowledge Discovery and Data Mining* (pp. 787-795).

[17] Shang, J., Ma, T., Xiao, C., & Sun, J. (2019). Pre-training of graph augmented transformers for medication recommendation. In *Proceedings of the 28th International Joint Conference on Artificial Intelligence (IJCAI)*, 5953-5959. https://doi.org/10.24963/ijcai.2019/825.

[18] Liu, Z., Li, X., Peng, H., He, L., & Yu, P. S. (2020, December). Heterogeneous similarity graph neural network on electronic health records. In *2020 IEEE International Conference on Big Data (Big Data)* (pp. 1196-1205). IEEE.

[19] Lu, H., & Uddin, S. (2023). Disease prediction using graph machine learning based on electronic health data: a review of approaches and trends. *Healthcare, 11*(7), 1031. https://doi.org/10.3390/healthcare11071031.

[20] Kipf, T. N., & Welling, M. (2017). Semi-supervised classification with graph convolutional networks. In *International Conference on Learning Representations (ICLR)*.

[21] Hamilton, W., Ying, Z., & Leskovec, J. (2017). Inductive representation learning on large graphs. *Advances in Neural Information Processing Systems*, 30.

[22] Veličković, P., Cucurull, G., Casanova, A., Romero, A., Liò, P., & Bengio, Y. (2018). Graph attention networks. In *International Conference on Learning Representations (ICLR)*.

[23] Cox, D. R. (1972). Regression models and life-tables. *Journal of the Royal Statistical Society: Series B (Methodological), 34*(2), 187-202.

[24] Katzman, J. L., Shaham, U., Cloninger, A., Bates, J., Jiang, T., & Kluger, Y. (2018). DeepSurv: personalized treatment recommender system using a Cox proportional hazards deep neural network. *BMC Medical Research Methodology, 18*(1), 24. https://doi.org/10.1186/s12874-018-0482-1.

[25] Lee, C., Zame, W. R., Yoon, J., & van der Schaar, M. (2018). DeepHit: a deep learning approach to survival analysis with competing risks. *Proceedings of the AAAI Conference on Artificial Intelligence, 32*(1), 2314-2321. https://doi.org/10.1609/aaai.v32i1.11842.

[26] Lee, C., Yoon, J., & van der Schaar, M. (2020). Dynamic-DeepHit: a deep learning approach for dynamic survival analysis with competing risks based on longitudinal data. *IEEE Transactions on Biomedical Engineering, 67*(1), 122-133. https://doi.org/10.1109/TBME.2019.2909027.

[27] Johnson, A., Bulgarelli, L., Pollard, T., Gow, B., Moody, B., Horng, S., Celi, L. A., & Mark, R. (2024). *MIMIC-IV (version 3.1)*. PhysioNet. RRID:SCR_007345. https://doi.org/10.13026/kpb9-mt58.

[28] Johnson, A. E. W., Bulgarelli, L., Shen, L., Gayles, A., Shammout, A., Horng, S., ... & Mark, R. G. (2023). MIMIC-IV, a freely accessible electronic health record dataset. *Scientific Data, 10*, 1. https://doi.org/10.1038/s41597-022-01899-x.

[29] Gu, A., & Dao, T. (2024). Mamba: linear-time sequence modeling with selective state spaces. In *Conference on Language Modeling (COLM)*. https://arxiv.org/abs/2312.00752.